%% file: full_paper_v1.tex
\documentclass[lettersize,journal]{IEEEtran}
\usepackage{amsmath,amsfonts}
\usepackage{algorithmic}
\usepackage{algorithm}
\usepackage{array}
\usepackage[caption=false,font=normalsize,labelfont=sf,textfont=sf]{subfig}
\usepackage{textcomp}
\usepackage{stfloats}
\usepackage{url}
\usepackage{verbatim}
\usepackage{graphicx}
\usepackage[backend=biber,style=ieee]{biblatex}
\usepackage{pifont}
\usepackage{multirow}
\usepackage{svg}
\usepackage{hyperref} 
\usepackage{tabularx}

\usepackage{authblk}

\newcommand{\rz}[1]{{\color{black}{#1}}}

\begin{document}

\title{GeoPix: Multi-Modal Large Language Model for Pixel-level Image Understanding in Remote Sensing}

\author{Ruizhe Ou, Yuan Hu$^*$, Fan Zhang, Jiaxin Chen, Yu Liu
\thanks{The work was supported by the National Natural Science Foundation
of China under Grant No. 42401407, No. 42430106, No. 42371468 and No. U24B20177, the China Postdoctoral Science Foundation under Grant No. 2024M760086, and the High-performance Computing Platform of Peking University No. HPC2306190166.}
\thanks{Ruizhe Ou and Jiaxin Chen are with the Pattern Recognition and Intelligent System Lab, School of Artificial Intelligence, Beijing University of Posts and Telecommunications, Beijing 100876, China (email: ouruizhe@bupt.edu.cn; chen\_jiaxin@bupt.edu.cn).}
\thanks{Yuan Hu, Fan Zhang and Yu Liu are with the Institute of Remote Sensing and Geographic Information Systems, School of Earth and Space Sciences, Peking University, Beijing 100871, China. Yu Liu is also with Peking University Ordos Research Institute of Energy, Ordos 017000, China (email: huyuan@pku.edu.cn; fanzhanggis@pku.edu.cn; yuliugis@pku.edu.cn).}
\thanks{$^*$ Corresponding author: Yuan Hu.}}



\maketitle

\begin{abstract}
Multi-modal large language models (MLLMs) have achieved remarkable success in image- and region-level remote sensing (RS) image understanding tasks, such as image captioning, visual question answering, and visual grounding. However, existing RS MLLMs lack the pixel-level dialogue capability, which involves responding to user instructions with segmentation masks for specific instances. In this paper, we propose GeoPix, a RS MLLM that extends image understanding capabilities to the pixel level. This is achieved by equipping the MLLM with a mask predictor, which transforms visual features from the vision encoder into masks conditioned on the LLM’s segmentation token embeddings. To facilitate the segmentation of multi-scale objects in RS imagery, a class-wise learnable memory module is integrated into the mask predictor to capture and store class-wise geo-context at the instance level across the entire dataset. In addition, to address the absence of large-scale datasets for training pixel-level RS MLLMs, we construct the GeoPixInstruct dataset, comprising 65,463 images and 140,412 instances, with each instance annotated with text descriptions, bounding boxes, and masks. Furthermore, we develop a two-stage training strategy to balance the distinct requirements of text generation and masks prediction in multi-modal multi-task optimization. Extensive experiments verify the effectiveness and superiority of GeoPix in pixel-level segmentation tasks, while also maintaining competitive performance in image- and region-level benchmarks. The models, dataset, and code are publicly available at \href{https://github.com/Norman-Ou/GeoPix}{https://github.com/Norman-Ou/GeoPix}.
\end{abstract}


\begin{figure}[t]
  \centering
  \includegraphics[width=\linewidth]{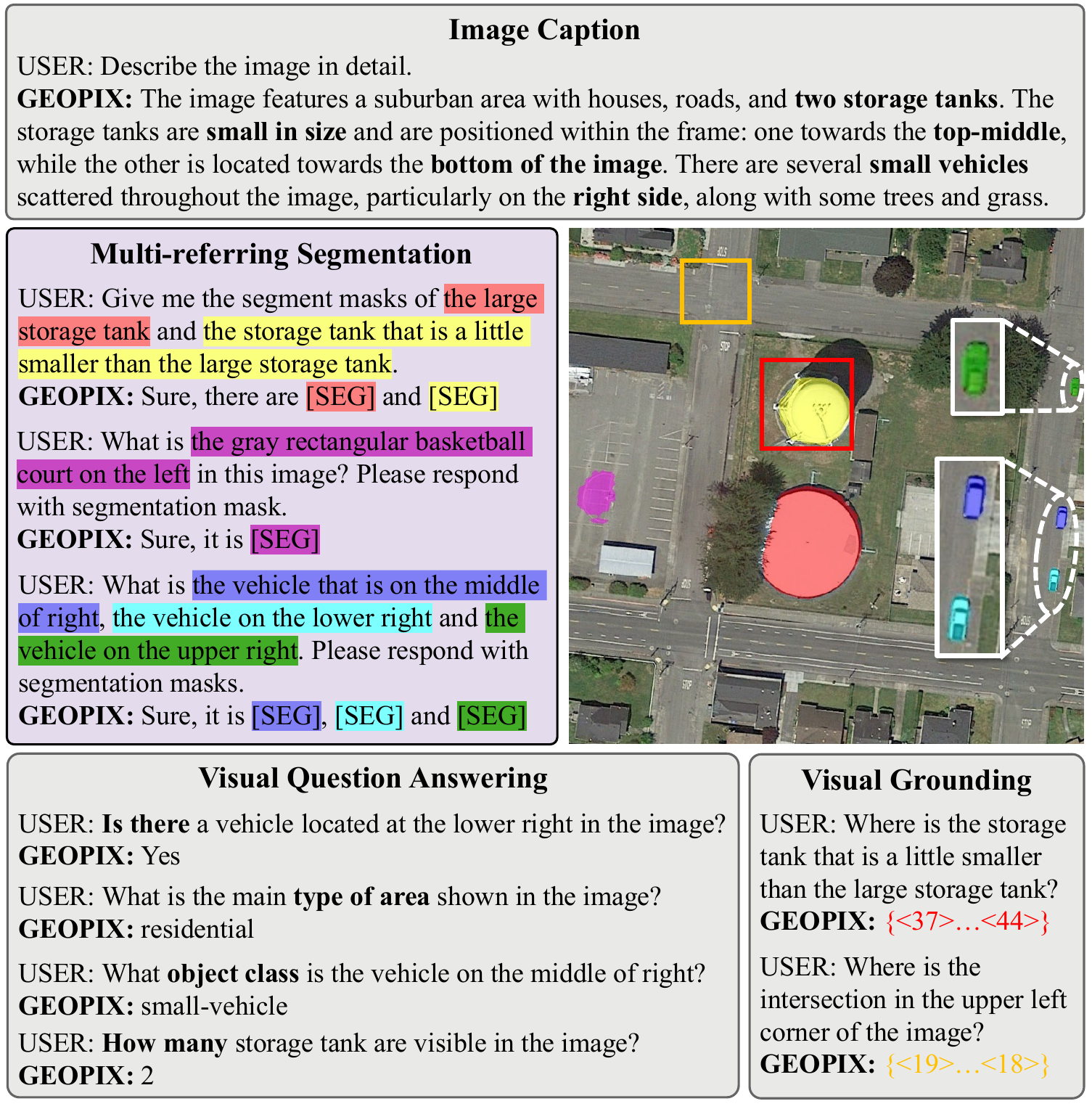}
  \caption{Overview of GeoPix capabilities in multi-modal, multi-task remote sensing image interpretation through dialogue.
  GeoPix supports image-level tasks, including image captioning and visual question answering, as well as region-level tasks like visual grounding.
  Additionally, it expands its functionality to pixel-level tasks, specifically multi-referring segmentation, as depicted in the purple panel.}
  \label{fig:intro}
\end{figure}

\section{Introduction}
\IEEEPARstart{M}{ulti-modal} large language models (MLLMs) in remote sensing (RS) have demonstrated outstanding performance on multi-modal tasks such as image captioning, visual question answering (VQA), and visual grounding.
For example, RSGPT~\cite{hu2023rsgpt} adapted a pre-trained MLLM to the RS domain, marking it as the first model to support image-level multi-modal tasks such as image captioning and image-level question answering.
Subsequent RS MLLMs~\cite{kuckreja2023geochat,skyeyegpt,earthgpt,BB-GeoGPT,Lhrs-bot,MTP,skysensegpt,teochat}, such as GeoChat~\cite{kuckreja2023geochat}, SkyEyeGPT~\cite{skyeyegpt}, \rz{EarthGPT~\cite{earthgpt}, LHRS-Bot~\cite{Lhrs-bot}, and EarthMarker~\cite{zhang2024earthmarker}}, have advanced to include region-level dialogue capabilities. These models support tasks such as region captioning, which involves generating captions for specific image regions based on user instructions, and visual grounding, where bounding boxes are predicted for instances in response to user prompts. 
However, existing RS MLLMs lack pixel-level dialogue capabilities, namely, the ability to generate pixel-level segmentation masks for specific instances based on user instructions.

To this end, we introduce GeoPix in this study, a RS MLLM that extends dialogue capabilities from image- and region-level to pixel-level image understanding.
Specifically, GeoPix supports image-level tasks including image captioning and visual question answering, as well as region-level tasks like visual grounding. Moreover, it extends support to pixel-level tasks, including referring segmentation and multi-referring segmentation.
Notably, we devise a Class-wise Learnable Memory (CLM) module to facilitate instance segmentation.
\rz{The CLM module adaptively learn the class-wise, instance-level shared geo-context information in remote sensing imagery.
The core component of the CLM module is a memory bank that learns and stores the shared information across the entire dataset during training.
During inference, the CLM retrieves the learned information from the memory bank. These learned high-level semantic information act as an adjuvant to improve the representation of instances, thereby facilitating segmentation.
}

In addition, the remote sensing field suffers from a lack of large-scale multi-modal datasets that simultaneously provide instance-level text descriptions, bounding boxes, and segmentation masks, which hampers the development of pixel-level RS MLLMs.
To address this gap, we constructed the GeoPixInstruct dataset, based on RSVGD~\cite{rsvgd} and SAMRS~\cite{SAMRS}. GeoPixInstruct is a comprehensive large-scale remote sensing multi-modal dataset encompassing 65,463 images and 140,412 instances.
Each instance is annotated with pixel-level masks, bounding boxes, and text descriptions, providing the support of pixel-level multi-modal tasks, such as multi-referring segmentation. 
Generating accurate descriptions for instances in remote sensing imagery is often challenging due to the varying spatial arrangements of objects, including isolated instances, linear arrangements, and clustered groupings. To address this, we develop a description generation pipeline based on GPT-4o, which identifies the spatial arrangement of instances and generates instance-level descriptions using provided examples for each arrangement type. To further enhance the accuracy of the generated descriptions, we manually verify a portion of the data and use it for vision fine-tuning of GPT-4o.
Tab.~\ref{tab:dataset_compare} provides a comparison between GeoPixInstruct and existing remote sensing multi-modal datasets.

In the optimization of pixel-level multi-task MLLMs, a tug-of-war problem arises: text generation typically requires fewer training steps to converge effectively, whereas segmentation demands extended training to achieve precise mask predictions. To address this, we introduce a two-stage training strategy that balances the distinct requirements of text generation and pixel-level mask prediction.
Specifically, in the first stage, our training focuses on equipping the model with text generation capabilities in the RS domain using fewer training steps, while introducing segmentation tokens to provide the model with initial mask prediction abilities. In the second stage, our training shifts its focus toward enhancing mask prediction accuracy by increasing the proportion of pixel-level training data and extending the training duration.
The two-stage strategy allows GeoPix to achieve state-of-the-art performance in multi-referring segmentation, while also delivering competitive or superior performance on image- and region-level multi-modal benchmarks.
Our contributions can be summarized as follows:
\begin{itemize}
    \item We propose GeoPix, a pixel-level RS MLLM that simultaneously supports image-, region- and pixel-level image understanding. 
    
    \item We devise a class-wise learnable memory module to effectively enhance segmentation accuracy and consistency of multi-scale objects in RS imagery.

    \item We construct the GeoPixInstruct dataset, which is a large-scale remote sensing multi-modal dataset that provides text descriptions, bounding boxes, and pixel-level masks for each instance.
    
    \item We develop a two-stage training strategy to manage the conflicting training requirements of text generation and mask prediction during the optimization.
\end{itemize}

\input{tables/intro_compare_data}

\section{Related Works}
\subsection{Multi-modal Large Language Models}
The typical architecture of existing multi-modal large language models (MLLMs) typically integrates a pre-trained multi-modal vision encoder with a large language model (LLM) via a linear layer or a MLP. This integration is achieved by concatenating visual tokens and text tokens, enabling the LLM to process multi-modal information simultaneously. 
Models such as LLaVA~\cite{liu2023llava,liu2023improvedllava,liu2024llavanext}, MiniGPT4~\cite{zhu2023minigpt,chen2023minigptv2}, Qwen-VL~\cite{bai2023qwenvlversatilevisionlanguagemodel}, and others~\cite{wang2023cogvlm,instructblip,otter,peng2023instructiontuninggpt4,interngpt,llamaadaptere,llamaadapterv2} leverage this approach to maintain strong language capabilities of LLM while supporting image-level and region-level multi-modal tasks, including image captioning, visual question answering, and visual grounding.
However, these models lack the capability for pixel-level image understanding. Subsequent models have addressed this issue; for example,
\rz{LISA~\cite{lisa} achieves referring segmentation during dialogue by integrating SAM~\cite{sam} with an LLM, while PixelLM~\cite{pixellm} accomplishes the same by redesigning a multi-scale mask decoder to accelerate inference and improve segmentation performance.}
These models establish a separate visual pathway to predict pixel-level masks conditioned on the output embeddings of LLM.
Overall, these approaches extend the MLLMs' capability to handle more fine-grained visual tasks, such as pixel-level segmentation.

\subsection{Remote Sensing Multi-modal Large Language Models}
\rz{Existing remote sensing (RS) MLLMs have demonstrated capabilities in image-, region- and pixel-level image understanding, enabling tasks such as image captioning, visual question answering, visual grounding and grounded conversation generation. }
For instance, RSGPT~\cite{hu2023rsgpt} adapt a pre-trained MLLM to the RS domain, supporting image-level tasks including image captioning and image-level question answering.
Models like GeoChat~\cite{kuckreja2023geochat}, SkyEyeGPT~\cite{skyeyegpt}, \rz{EarthGPT~\cite{earthgpt}, Lhrs-bot~\cite{Lhrs-bot}, EarthMarker~\cite{zhang2024earthmarker}, and others~\cite{BB-GeoGPT, MTP, skysensegpt, teochat}} extended these capabilities to the region-level, supporting tasks like grounded image captioning, region-specific visual question answering, and visual grounding.

\rz{While the aforementioned RS MLLMs have achieved image-level and region-level dialogue capabilities in the RS domain, pixel-level RS MLLMs have been attempted only to a limited extent. }
\rz{Pixel-level vision-language capabilities were first introduced by RRSIS~\cite{yuan2024rrsis} and RMSIN~\cite{liu2024rmsin}. These models are limited to single-instance referring image segmentation and lack a broader vision-language interaction. 
Concurrent work, GeoPixel~\cite{shabbir2025geopixelpixelgroundinglarge}, aimed to introduce pixel-level dialogue capabilities into the remote sensing (RS) domain by leveraging the LISA~\cite{lisa} architecture. Our work differs from GeoPixel in the range of supported tasks, model architecture, dataset construction methodology, and training strategy.}

\subsection{Remote Sensing Multi-modal Datasets}
In the field of RS, existing open-source multi-modal datasets only support image-level and region-level multi-modal tasks. For example, RSICap~\cite{hu2023rsgpt} contains only image-text pairs and facilitates image-level tasks such as image captioning and visual question answering.
GeoChatInstruct~\cite{kuckreja2023geochat}, RSVP~\cite{zhang2024earthmarker}, MMRS-1M~\cite{earthgpt}, LHRS-Bench~\cite{Lhrs-bot}, and VRSBench~\cite{li2024vrsbench} are build on existing datasets~\cite{SAMRS,DIOR,dota,fair1m,rsvgd,Xia_2017,gao2020nwpu,helber2019eurosat,Lobry_2020} by automatically generating text descriptions for objects or regions within images using rule-based methods or GPT models, enabling region-level tasks like grounded image captioning, region-specific question answering, and visual grounding.
However, datasets with pixel-level image-text labels, such as RefSegRS~\cite{yuan2024rrsis}, RRSIS-D~\cite{liu2024rmsin} are designed for referring image segmentation tasks. They do not include bounding box labels, their annotations are not intended for instruction-following, and they are relatively small in scale.
Therefore, we constructed a large-scale dataset to support the training of pixel-level RS MLLMs.

\begin{figure*}[!t]
  \centering
  \includegraphics[width=\textwidth]{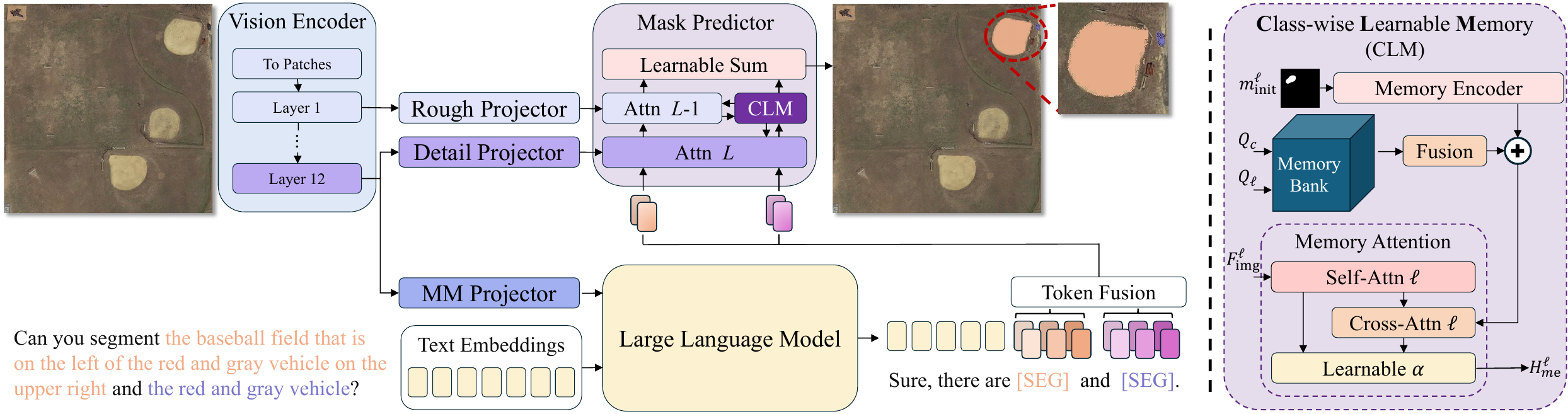}
  \caption{Overview of the proposed GeoPix model architecture and the detailed design of the class-wise learnable memory (CLM) module. The left panel illustrates the overall architecture of GeoPix.
  GeoPix takes user text instructions and images as input and outputs corresponding answers and segmentation results (if the user’s instruction requests single or multiple object segmentation).
  The user-input image is encoded to extract multi-scale visual features via the vision encoder. These features are transformed by independent projectors and fed into both the mask predictor and the LLM. The LLM processes the instructions along with the deepest scale of visual features to predict interleaved text and segmentation tokens. Subsequently, the segmentation tokens serve as conditions for the mask predictor, guiding it to predict masks for the instances specified by the user. The right panel showcases the detailed design of the CLM module. The CLM module first encodes the initial mask \( m_{\text{init}}^{\ell} \) of each scale \(\ell\) into latent representations using the memory encoder, and then retrieves memory features from the memory bank using the category query \( Q_{c} \) and scale query \( Q_{\ell} \). The memory features are combined with the encoded initial mask via element-wise addition. Finally, the aggregated features serve as the key and value in the memory attention module, enhancing the visual features to obtain memory-enhanced features \(H_\text{me}\).}
  \label{fig1}
\end{figure*}

\subsection{Memory Bank}
Memory banks have been widely employed in computer vision tasks such as classification~\cite{lanchantin2017memory, liu2018unsupervised, liu2024point}, detection~\cite{zhou2017new, xing2023visual, sun2021mamba}, and segmentation~\cite{memorybanksemi, du2022weakly, mcic}, enhancing performance by learning and storing shared features during training and integrating them into predictions for new inputs. In segmentation~\cite{mcic}, a class-wise memory module retains features from class-wise objects, supporting segmentation by leveraging intra-class consistency. Recently, SAM-2~\cite{sam2} applied a memory bank to video instance segmentation. The memory bank in SAM-2 stores instance-level features from previous frames and uses them as conditions for segmenting subsequent frames, ensuring segmentation consistency for the same object. Similarly, Skysense~\cite{skysense} introduces prototype learning in the remote sensing domain, employing a learnable approach to extract class-wise geo-context representations. By providing complementary cues, prototype learning enhances feature representations, resulting in improved performance in remote sensing tasks such as segmentation and detection. 
The difference between prototype learning and memory banks lies in that prototype learning employs clustering to identify class centers or prototypes, whereas memory-based methods focus on storing specific features of samples. In this study, we utilize the memory bank mechanism to store class-wise geo-context in remote sensing imagery, thus facilitating segmentation.

\section{Method}  
In this section, we first introduce the overall architecture of GeoPix, then elaborate on the proposed  class-wise learnable memory module, and finally, we describe the two-stage training strategy.

\subsection{GeoPix Architecture}  
GeoPix is composed of three primary components: i) a pre-trained vision encoder, ii) a large language model (LLM), iii) a multi-scale mask predictor with class-wise learnable memory.
The overall architecture is depicted in Fig.~\ref{fig1}. GeoPix processes the input image and user instructions and generates interleaved text descriptions and corresponding masks for a varying number of targets.

For an input image \( x_{\text{img}} \), the vision encoder \( \mathcal{E} \) extracts a spectrum of multi-scale visual features, represented as \( \{ F_{\text{img}}^{\ell} = \mathcal{E}(x_{\text{img}}) \}_{\ell=1}^{L} \), where \( L \) denotes the number of visual scales. Each visual feature \( F_{\text{img}}^{\ell} \) is further transformed by scale-specific projectors \( \mathcal{P}^{\ell} \), resulting in \( \{ F_{\text{img}}^{\ell} = \mathcal{P}^{\ell}(F_{\text{img}}^{\ell}) \}_{\ell=1}^{L} \). This forms separate visual pathways connected to the mask predictor. Additionally, the deepest visual feature \( F_{\text{img}}^{L} \) is processed through a multi-modal (MM) projector \( \mathcal{P}^{L} \), resulting in \( F_{\text{llm\_img}} = \mathcal{P}^{L}(F_{\text{img}}^{L}) \).
This output is then fed into the LLM to generate textual outputs, and segmentation tokens specifically intended to trigger the mask predictor.

Segmentation tokens are introduced into the LLM to serve as conditioning inputs following~\cite{pixellm}. The segmentation tokens are defined as 
\( T_{seg} = \{\tau_{n}^{\ell} \in \mathbb{R}^{d}\}^{N,L}_{n=1,\ell=1} \), 
where \( L \) represents the number of scales, \( N \) denotes the number of tokens per scale, and \(d\) is the hidden dimension of LLM. These tokens are assigned to each scale, providing scale-specific conditions for the segmentation of each instance, with their embeddings serving as conditioning inputs for the multi-scale mask predictor to facilitate the segmentation of each corresponding instance.

Conditioned by the embeddings of segmentation tokens, the mask predictor interacts with the proposed class-wise learnable memory (CLM) module to transform multi-scale visual features into precise masks. Specifically, the mask predictor first generates initial masks  \( M_{\text{init}} = \{m^{\ell}_{\text{init}} \in \mathbb{R}^{K \times H_{p} \times W_{p}}\}_{\ell=1}^{L} \), for each scale \(\ell\), where \( K \) denotes the number of targets, and \( H_{p} \) and \( W_{p} \) represent the height and width of the pixel-level masks. These initial masks are subsequently refined by the CLM module to produce more accurate segmentation masks. The final masks are obtained through a learnable summation of the masks generated from visual features at different scales, formulated as
\(P_\text{mask} = \sum^{L}_{\ell=1} \beta \cdot P^{\ell}_\text{mask}\), where \(\beta\) is learnable.
In the following, we detail the design of the proposed CLM module.

\subsection{Class-wise Learnable Memory (CLM) Module}
\label{sec:clm}
\rz{The remote sensing imagery typically features a vertical-downward capture angle, leading to instances of the same class sharing similar geo-contexts, such as shape, across different images. These geo-contexts remain relatively stable regardless of variations in ground sampling distance (GSD) or the region of acquisition. To adaptively learn this geo-context, we propose a Class-wise Learnable Memory (CLM) module that extracts and stores high-level semantic information of shared geo-contexts within each category. These high-level semantic information act as an adjuvant to improve the representation of specific instances, thereby facilitating segmentation.} 
The module consists of three components: i) a memory encoder, ii) a memory bank, and iii) a memory attention block, as depicted in the right panel of Fig.~\ref{fig1}. 

To adaptively store semantic information across different scales and categories, we design a matrix-like memory bank \(\mathcal{M}\) organized by class and scale indices. The scale of visual features and class information serve as queries to retrieve relevant memory features \( H_m \) from the memory bank, as shown in Eq.~\ref{eq:memory}.

\begin{equation}
H_m = \mathcal{M}(Q_{c}, \; Q_{\ell})
\label{eq:memory}
\end{equation}
\noindent where \( H_m = \{h^{\ell} \in \mathbb{R}^{K\times N \times D \times H \times W} \}_{\ell=1}^{L} \), \(K\) and \( L \) represent the number of classes and scales. \( N \) denotes the memory capacity for expanding the storage capability of each class, and \( D \) denotes the dimension of the memory, while \( H \) and \( W \) correspond to the height and width of the memory features.
$Q_{c}$ and $Q_{\ell}$ denote the category query and the scale query that are used to extract class-wise geo-context of specific scale.

Building on this memory bank, we introduce a memory encoder \( \mathcal{I} \) and a memory attention block to fuse the multi-scale visual features \( \{ F_{\text{img}}^{\ell} = \mathcal{P}^{\ell}(F_{\text{img}}^{\ell}) \}_{\ell=1}^{L} \) and initial masks \( m_{\text{init}}^{\ell} \), resulting in memory-enhanced features \( H_{\text{me}} \), as expressed in Eq.~\ref{eq:memen}:

\begin{equation}
H_{\text{me}} = \mathcal{CA}( \mathcal{SA}(F^{\ell}_{\text{img}}), \; \mathcal{I}(m_{\text{init}}^{\ell}) + \mathcal{F}(h^{\ell}))
\label{eq:memen}
\end{equation}
\noindent where \(\mathcal{CA}\) and \(\mathcal{SA}\) represent cross attention and self-attention, respectively. \(\mathcal{F}\) denotes the fusion of memory features. The memory feature fusion \(\mathcal{F}\) is performed along the memory capacity \( N \), formulated as \( \mathcal{F}(h^{\ell}) \in \mathbb{R}^{ K\times 1 \times D\times H\times W} \). 
\( \mathcal{I}(m_{\text{init}}^{\ell})\) represents the encoded pixel-level initial masks, where \( \mathcal{I}(m_{\text{init}}^{\ell}) \in \mathbb{R}^{K \times D\times H \times W}\). 
The fused memory features are then combined with the encoded initial mask via element-wise addition. 
Memory fusion enables the learnable extraction and integration of the most valuable information from the memory capacity.
The aggregated representation serves as the key and value in the \( \mathcal{CA} \), interacting with the self-attended visual features \( \mathcal{SA}(F^{\ell}_{\text{img}}) \) to generate the memory enhanced semantic features \( H_{\text{me}} \). The final output is further refined through a learnable parameter \( \alpha \), forming a residual connection with \( \mathcal{SA}(F^{\ell}_{\text{img}}) \) to ensure feature consistency and stability, as shown in Eq.~\ref{eq:residual}:

\begin{equation}
H_{\text{me}} = \mathcal{SA}(F^{\ell}_{\text{img}}) + \alpha \cdot H_{\text{me}}
\label{eq:residual}
\end{equation}

\noindent The memory-enhanced features enrich the visual representation at each scale by incorporating higher-level semantic information of class-wise shared geo-contexts.

\begin{figure*}[t]
  \centering
  \includegraphics[width=\textwidth]{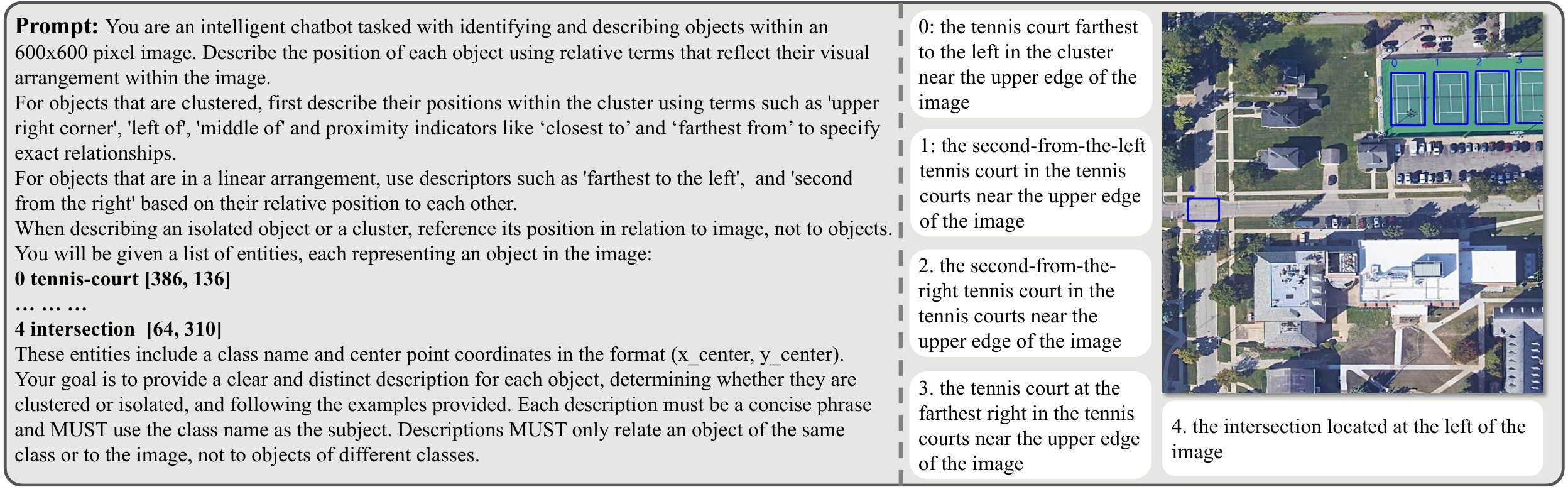}
  \caption{The left panel displays the instruction used to prompt both GPT-4o and its fine-tuned version within our description generation pipeline. The right panel provides an example description generated by the fine-tuned GPT-4o. Bounding boxes are shown for instance differentiation and are not part of GPT-4o’s input.}
  \label{fig:data}
\end{figure*}

\subsection{Two-stage Training Strategy}  
Balancing multi-task training for text generation and multi-referring segmentation presents significant challenges. Text generation tasks typically achieve convergence within fewer training epochs, whereas multi-referring segmentation demands more extensive training to ensure precision.
To tackle this disparity, we adopt a two-stage training strategy for GeoPix. In the first stage, the LLM is fine-tuned, while the mask predictor and the CLM module are initially trained from scratch. The model is trained for five epochs using the full VRSBench dataset in combination with GeoPixInstruct. This stage primarily focuses on adapting the pre-trained LLM’s image-level and region-level dialogue capabilities to the remote sensing domain, while simultaneously introducing segmentation tokens to facilitate initial mask prediction. In the second stage, the model undergoes further training with an emphasis on refining segmentation. At this stage, 8,000 samples are randomly selected from the VRSBench to prevent catastrophic forgetting of image-level and region-level dialogue capabilities and are combined with data from each subset of GeoPixInstruct. The proposed two-stage strategy enables progressive refinement of pixel-level segmentation performance without compromising the model’s ability to generate text. This approach ensures that the model excels in both segmentation and text generation tasks across multi-modal remote sensing applications.

\section{GeoPixInstruct: A Pixel-level Instruction Dataset for Remote Sensing}
In this section, we introduce the GeoPixInstruct dataset, a large-scale remote sensing dataset specifically designed to support the training of pixel-level RS MLLMs.
GeoPixInstruct is developed based on SAMRS~\cite{SAMRS}. As shown in Tab.~\ref{tab:dataset_compare}, SAMRS comprises three subsets—SIOR, FAST, and SOTA—where each instance is annotated with bounding box and mask. 
We supplemented the SAMRS dataset with instance-level textual descriptions. Consequently, GeoPixInstruct also includes three subsets: SIOR-T (derived from SIOR), FAST-T (derived from FAST), and SOTA-T (derived from SOTA).

\subsection{Systematic Data Filtering}
When constructing the textual descriptions for each instance, our goal is to ensure that the text uniquely refers to the corresponding instance.
However, remote sensing images often contain densely clustered instances of the same category, such as yacht marinas or parking lots, posing challenges for generating precise textual descriptions for each instance for such images.
Therefore, we first apply predefined rules, as referenced in~\cite{rsvgd}, to filter images from the existing instance segmentation dataset. Specifically, we ensure that the number of instances of the same category within a single image remains below five, and the total number of instances across different categories does not exceed eleven. This filtering process yields 65,463 images and 140,412 instances, encompassing both isolated and clustered instances.

\subsection{Construction of SIOR-T}
The SIOR subset of SAMRS is constructed based on DIOR~\cite{DIOR} and shares the same images as RSVG-DIOR (RSVGD)~\cite{rsvgd}. Since RSVGD provides instance-level descriptions, we integrate these descriptions with the annotations from SIOR to form SIOR-T by matching instance bounding box coordinates between the two datasets, allowing for a margin of error of 10 pixels. This integration equips SIOR-T with both instance-level descriptions and pixel-level mask annotations.
In contrast, the other two subsets of SAMRS—FAST and SOTA—do not have existing open-source descriptions. To address this limitation, we designed an automatic description generation pipeline to produce instance-level descriptions for FAST and SOTA, thereby constructing the FAST-T and SOTA-T subsets.


\subsection{Construction of FAST-T and SOTA-T}
Previous works\cite{li2024vrsbench, lisa, pixellm} have demonstrated the feasibility of constructing descriptions using GPT. Building upon this, we designed a two-step description generation pipeline leveraging GPT-4o. In the first step, we utilize a structured prompt to guide GPT-4o in generating initial descriptions in a zero-shot manner. A subset of these initial descriptions is then manually curated to form a fine-tuning dataset. This dataset is used for vision tuning GPT-4o, enhancing the accuracy of the generated instance-level descriptions. 
In the structured prompt, we guide GPT-4o to first determine the spatial arrangement status of objects, categorizing them as isolated, arranged in a line, or clustered.
For isolated instances, the descriptions follow the methodology outlined in RSVGD, highlighting the attributes of the instance and its spatial relationships relative to other instances.
For linear and clustered instances, the descriptions begin with the relative position of the instance within the line or cluster, followed by the line or cluster’s overall position in the image. The prompt also includes a detailed list specifying each instance’s index, category, and center point coordinates to ensure comprehensive and accurate instructions for GPT-4o.
An example of the structured prompt is shown in Fig.~\ref{fig:data}. 

In the second step, to improve the accuracy of descriptions generated by GPT-4o, we implemented a refinement process involving human feedback and fine-tuning. Specifically, 100 images containing isolated instances, instances in linear arrangements, and clustered instances were randomly selected from each category for manual correction of the initial descriptions generated by GPT-4o.
Descriptions exhibiting inconsistencies—particularly for clustered instances—were revised to address noise in the generated outputs. These corrected descriptions formed a high-quality fine-tuning dataset, which was subsequently used to conduct vision-based fine-tuning on GPT-4o. This iterative process enhanced the model’s ability to produce accurate and precise descriptions for instances across diverse spatial configurations, improving performance for both isolated and clustered instances.

\begin{figure*}[t]
  \centering
  \includegraphics[width=\linewidth]{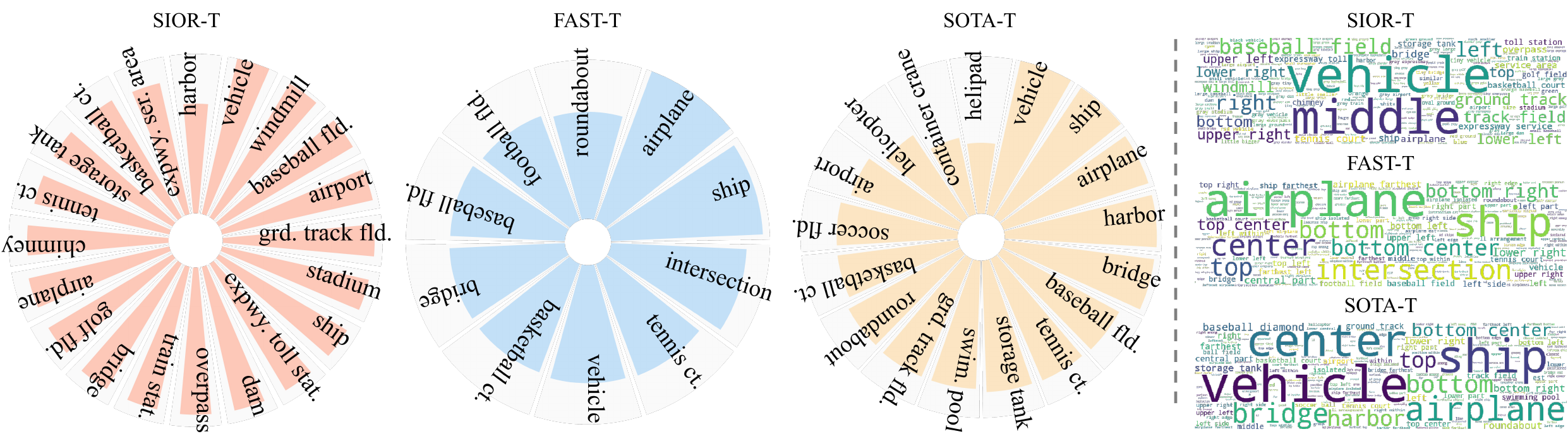}
  \caption{\rz{Category distribution analysis and word cloud visualization for the SIOR-T, FAST-T, and SOTA-T subsets of GeoPixInstruct. ``expwy. ser. area'' stands for expressway service area, ``expwy. toll stat.'' stands for expressway toll station, ``grd. track fld.'' stands for ground track field, ``stat.'' is the abbreviation for station, ``ct.'' is the abbreviation for court, and ``fld.'' is the abbreviation for field.}}
  \label{fig:wordcloud}
\end{figure*}

\subsection{Statistics}
\begin{figure}[t]
  \centering
  \includegraphics[width=\linewidth]{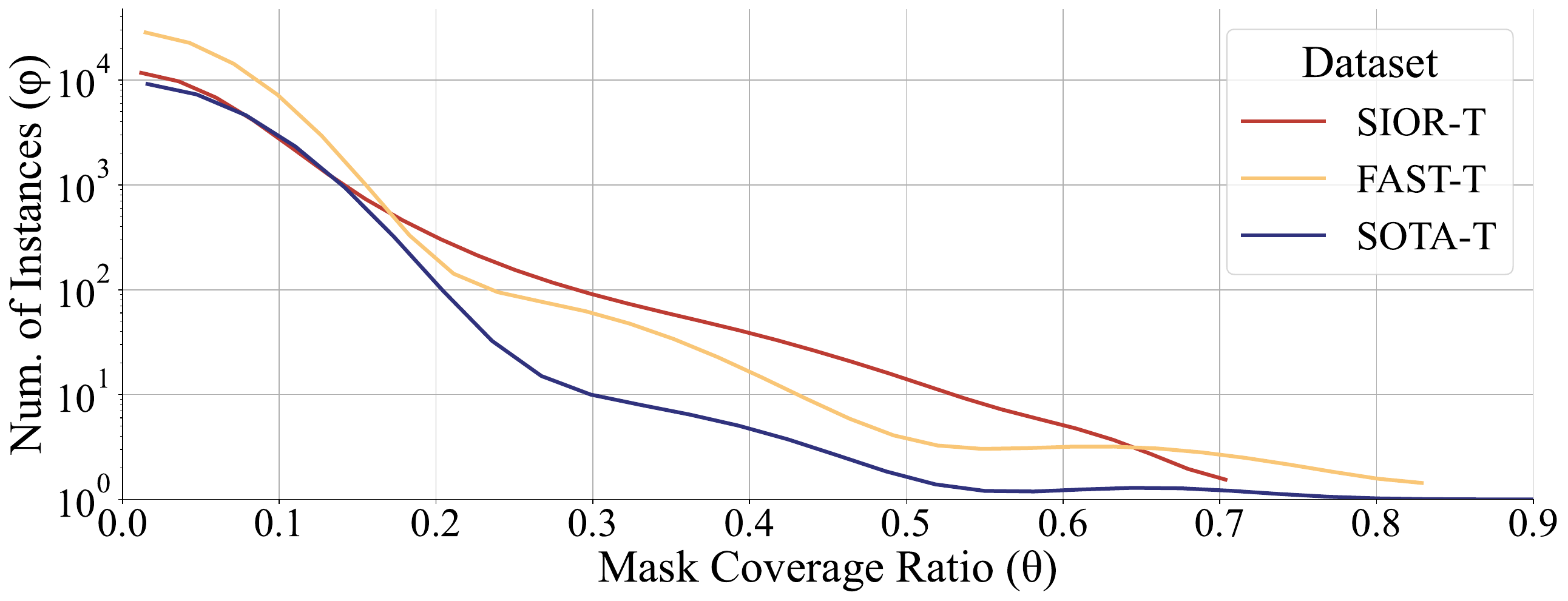}
  \caption{The distribution of instance number (\(\varphi\)) by mask coverage ratio (\(\theta\)). Subset located toward the lower-left corner represent smaller instance number with lower coverage ratio, indicating higher segmentation difficulty.}
  \label{fig:ratio}
\end{figure}

GeoPixInstruct comprises three subsets—SIOR-T, FAST-T, and SOTA-T—containing a total of 65,463 images and 140,412 instances. Each instance is annotated with a textual description, instance category, bounding box, and mask. 
\rz{Fig.~\ref{fig:wordcloud} presents the word cloud visualization and category distribution for the three subsets, illustrating their textual and categorical characteristics.} 
As shown in Tab.~\ref{table:data}, we provide detailed statistics on the number of images, categories, instances, and corresponding annotations for each subset, along with the image size \rz{and ground sampling distance (GSD) in meters}. 
Additionally, we report the average number of instances per image (\(\varphi\)) and the average mask coverage ratio (\(\theta\)) for each subset. \(\theta\) is calculated as the ratio of the pixel area of an instance mask to the total pixel area of the image. 
To further analyze the each subset's characteristics, we visualize the distribution of instance counts by mask coverage ratio (\(\theta\)) in Fig.~\ref{fig:ratio}. The three subsets exhibit distinct variations in \(\varphi\) and \(\theta\), with their relationship reflecting differing segmentation difficulty levels. Specifically, lower \(\theta\) and fewer instances correspond to increased segmentation difficulty. Based on this, the SOTA-T subset is identified as the most challenging, followed by FAST-T, with SIOR-T presenting the lowest level of difficulty.  

\input{tables/data}

\input{tables/seg_v1}

\begin{figure*}[t]
  \centering
  \includegraphics[width=\textwidth]{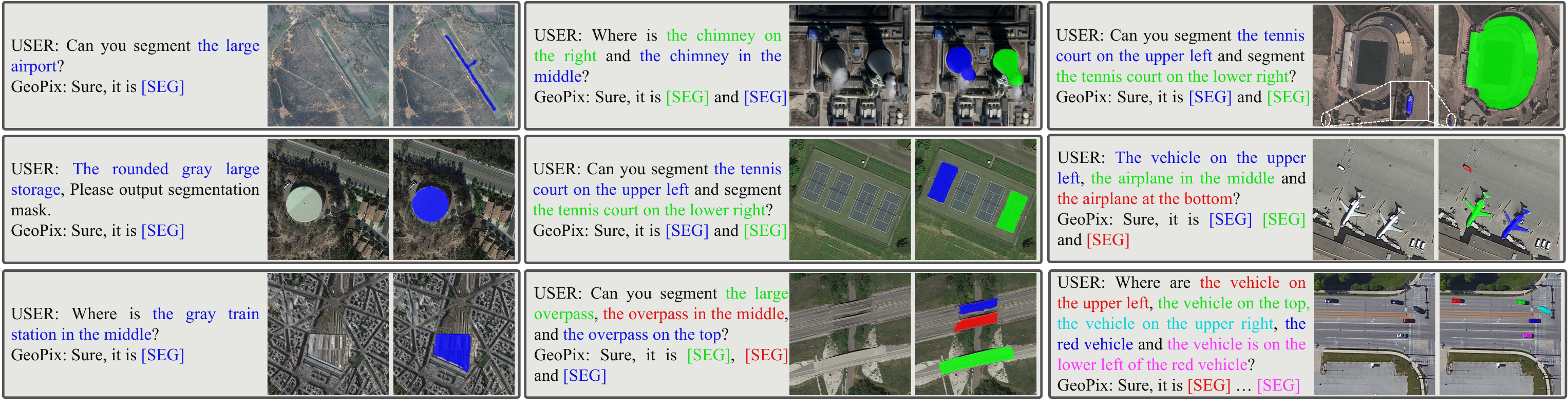}
  \caption{Visualization of GeoPix referring segmentation results. The left panel displays single-instance referring segmentation. The middle panel showcases multi-instance referring segmentation for instances within the same category. The right panel illustrates multi-instance referring segmentation across different categories.}
  \label{fig:seg_text_result}
\end{figure*}

\section{Experiments}
\subsection{Implementation Details}
The vision encoder, MM Projector, and LLM in our model are initialized using the pre-trained LLaVA-1.5 model, while the remaining modules are trained from scratch. The model is consistently trained at an image resolution of \(336 \times 336\). In our implementation, the vision encoder remains frozen, and LoRA~\cite{lora} is applied to finetune the \( W_q \) and \( W_v \) parameters of LLM.
\rz{During stage one of our two-stage training strategy, the LoRA rank \( r \) is set to 128. This value is chosen empirically based on preliminary experiments aimed at increasing the trainable parameters of the LLM to adapt its capabilities from the general domain to the RS domain. In stage two, the rank \( r \) is reduced to 8. This change is driven by the need for fine-tuning, where a smaller rank enables the LLM to focus on predicting more accurate segmentation token representations, thereby improving segmentation performance while avoiding catastrophic forgetting of other multi-modal tasks.}
We adopt the AdamW optimizer, employing a linear warm-up phase followed by a cosine annealing learning rate scheduler. In stage one, we train for a total of 5 epochs, with the learning rate linearly increasing to \(3 \times 10^{-4}\) during the first epoch as a warm-up. In stage two, the learning rate follows the same warm-up schedule over 2 epochs, with training extended to 50 epochs to ensure optimal performance. The global batch size is maintained at 196 throughout the training period.

\subsection{Multi-referring Segmentation}
\label{sec:segmentation}
We first evaluate the performance of our model on the multi-referring segmentation task. 
We compare GeoPix with RRSIS~\cite{yuan2024rrsis}, RMSIN~\cite{liu2024rmsin}, LISA-7B~\cite{lisa}, and PixelLM-7B~\cite{pixellm} on GeoPixInstruct. 
RRSIS and RMSIN are specialized referring segmentation models in the remote sensing (RS) domain, while LISA and PixelLM are multi-modal large language models (MLLMs) in the general domain. 
All models are trained on the GeoPixInstruct dataset using their original training settings. \rz{For LISA and PixelLM, we additionally train them with the proposed two-stage training strategy.}
RRSIS and RMSIN can only predict one mask at a time based on a given textual description. Therefore, we enable RRSIS and RMSIN to perform single-instance referring segmentation multiple times to accomplish multi-referring segmentation.
In contrast, GeoPix, LISA, and PixelLM can directly perform multi-referring segmentation by predicting masks for multiple instances simultaneously.

\rz{Moreover, we group the instances with different size based on their mask coverage ratios (\(\theta\)) within the image. Specifically, we differentiate small, medium, and large instances by taking the logarithm of \(\theta\), where small corresponds to 0\textless\(\theta\)\textless0.01, medium corresponds to 0.01\textless\(\theta\)\textless0.1, and large corresponds to 0.1\textless\(\theta\)\textless1. }
As shown in Tab.~\ref{table:seg}, GeoPix outperforms the specialized models RRSIS and RMSIN on the SIOR-T subset while delivering competitive performance on the FAST-T and SOTA-T subsets. Additionally, GeoPix surpasses LISA and PixelLM across all subsets of GeoPixInstruct.

\begin{figure*}[t]
  \centering
  \includegraphics[width=\textwidth]{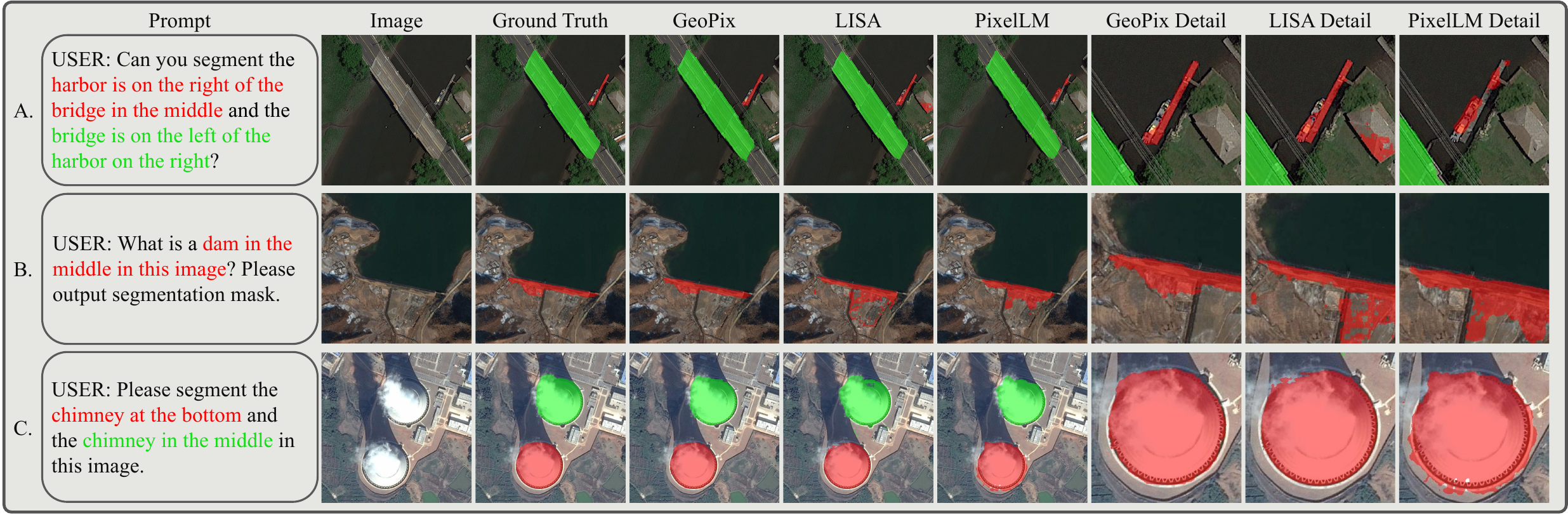}
  \caption{\rz{Visual comparison of segmentation results between GeoPix, LISA, and PixelLM. Specific regions of the original image are zoomed in to highlight segmentation differences. In Row A, GeoPix accurately segments the harbor, while LISA erroneously includes additional similar regions, and PixelLM produces a fragmented segmentation. In Row B, under the geographical similarity of the dam and its surrounding environment, GeoPix still accurately segments the dam from the surrounding environment, while LISA and PixelLM incorrectly segment the surrounding environment. In Row C, GeoPix produces a segmentation with smoother boundaries compared to both LISA and PixelLM.}}
  \label{fig:seg_compared_lisa_pixellm}
\end{figure*}

\begin{figure*}[t]
  \centering
  \includegraphics[width=\textwidth]{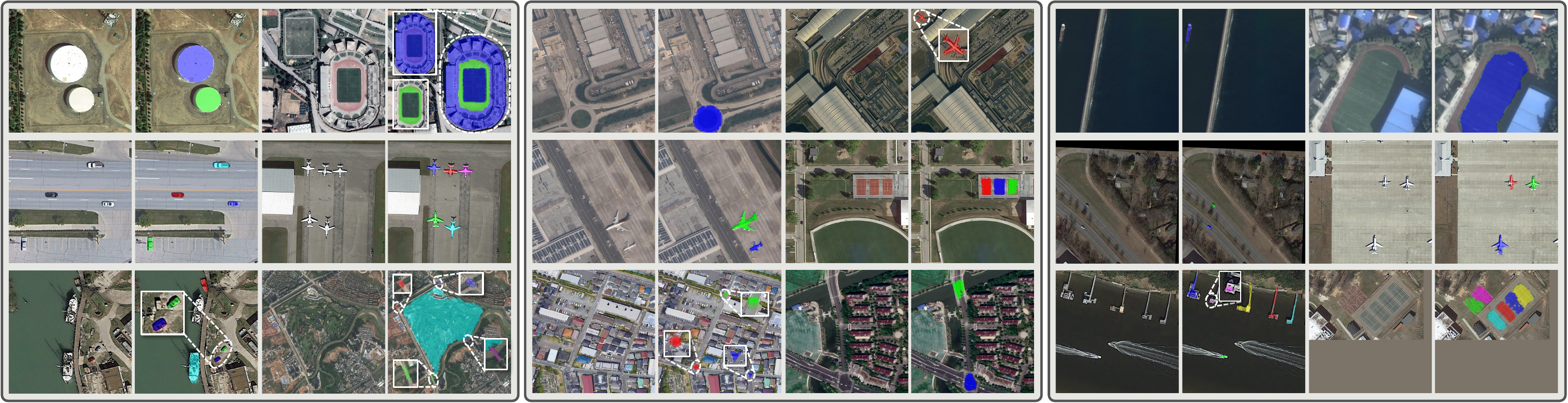}
  \caption{Visualization of additional GeoPix segmentation results across the three subsets of GeoPixInstruct. The left panel represents the SIOR-T subset, the middle panel corresponds to the FAST-T subset, and the right panel depicts the SOTA-T subset. Each panel presents examples in order from top to bottom, showcasing results for large to small instances, increasing instance counts, and transitions from single to multiple categories.}
  \label{fig:seg_result}
\end{figure*}

Fig.~\ref{fig:seg_text_result} displays examples of GeoPix’s dialogues for multi-referring segmentation, showcasing segmentation results for a single instance, instances within the same category, and instances across multiple categories. The results show that GeoPix can accurately locate objects specified by users and produce the corresponding segmentation masks. 

\rz{Fig.~\ref{fig:seg_compared_lisa_pixellm} visualizes the segmentation results of GeoPix, LISA, and PixelLM. GeoPix outperforms the other models in segmenting. In Row A, GeoPix successfully segments the harbor as a whole, while LISA erroneously includes additional similar regions, and PixelLM generates a fragmented segmentation. In Row B, under the geographical similarity of the dam and its surrounding environment, GeoPix still accurately segments the dam from the surrounding environment, while LISA and PixelLM incorrectly segment the surrounding environment. In Row C, the boundaries of the masks predicted by GeoPix are smoother and more complete compared to LISA and PixelLM.} 
In Fig.~\ref{fig:seg_result}, we present additional segmentation results from GeoPix across the three subsets of GeoPixInstruct. \rz{These results encompass segmentation outcomes on the same remote sensing imagery, featuring varying instance counts, sizes, and categories. They demonstrate that GeoPix consistently maintains accurate segmentation capabilities across remote sensing imagery with varying instance counts, sizes, and categories.}

\rz{Additionally, we evaluate the segmentation performance of applying SAM~\cite{sam} and SAM2~\cite{sam2} to the bounding boxes generated by visual grounding on GeoPixInstruct. GeoPix initially predicts the bounding boxes of the instances specified in the prompt, after which SAM or SAM2 leverages the predicted bounding boxes from GeoPix to generate the corresponding instance masks. As shown in Tab.~\ref{table:seg}, directly predicting the masks of instances using GeoPix results in better performance than first predicting the bounding boxes and then predicting the masks. This is likely due to the noise present in the bounding boxes predicted in a text generation manner using LLMs, which affects the segmentation performance of SAM and SAM2. Further evaluation on visual grounding can be found in Sec.~\ref{sec:visual_grounding}.}

\input{tables/vqa_vrsbench}
\input{tables/vqa_rsieval}
\input{tables/cap_rsieval}
\input{tables/cap_vrsbench}

\subsection{Visual Question Answering}
We evaluate the visual question answering (VQA) performance of GeoPix on VRSBench~\cite{li2024vrsbench} and RSIEval~\cite{hu2023rsgpt}.
On VRSBench, we compare GeoPix with MiniGPT4-v2~\cite{chen2023minigptv2}, LLaVA-1.5~\cite{liu2023improvedllava}, Mini-Gemini\cite{minigemini}, and GeoChat~\cite{kuckreja2023geochat}, all of which are fine-tuned on VRSBench. On RSIEval, GeoPix is compared with BLIP2, MiniGPT4~\cite{zhu2023minigpt}, and RSGPT~\cite{hu2023rsgpt}, where GeoPix performs predictions in a zero-shot manner.
VRSBench consists of 9,349 images paired with 37,409 question-answer pairs, encompassing ten question types: object category, presence, quantity, color, shape, size, position, direction, scene characteristics, and reasoning. RSIEval includes 100 images with 936 question-answer pairs, covering ten categories: presence, quantity, color, absolute position, relative position, area comparison, road direction, image content, scene, and reasoning. Both benchmarks include object-level, scene-level, and reasoning-level questions, while RSIEval additionally incorporates image-level questions. To quantify model performance, we use GPT to evaluate whether predicted answers align with ground truth, calculating accuracy as the metric.
Tab.~\ref{table:vqa_vrsbench} presents the results on VRSBench.
GeoPix achieves superior performance across all question types, surpassing the previous RS MLLM GeoChat~\cite{kuckreja2023geochat}.
Specifically, GeoPix achieves gains over GeoChat, ranging from 1.6\% (presence) to 24.4\% (category), with an average overall accuracy increase of 14\%.
Similarly, Tab.~\ref{table:vqa_rsieval} presents the results on RSIEval. GeoPix demonstrates competitive or superior zero-shot performance compared to previous MLLMs, outperforming RSGPT~\cite{hu2023rsgpt} by 2\%-13\% in question types such as color and absolute area comparison.
Overall, GeoPix demonstrates superior VQA performance compared to previous models, highlighting its effectiveness in VQA for remote sensing imagery.

\subsection{Image Captioning}
We evaluate the image captioning performance of GeoPix on VRSBench~\cite{li2024vrsbench} and RSIEval~\cite{hu2023rsgpt}.
GeoPix is compared with the same models mentioned in the VQA section. Note that GeoPix is evaluated in a zero-shot manner on RSIEval.
The models are prompted with the instruction \texttt{"Describe the image in detail."} and the generated captions are evaluated using metrics including BLEU~\cite{bleu}, ROUGE\_L~\cite{lin2004rouge}, METEOR~\cite{METEOR}, CIDEr~\cite{CIDEr}, and CLAIR~\cite{chan2023clairevaluatingimagecaptions}. Tab.~\ref{table:cap_vrsbench} presents the results on VRSBench, where GeoPix demonstrates competitive performance compared with GeoChat. 
Tab.~\ref{table:cap_rsieval} shows the results on RSIEval, GeoPix underperforms compared to RSGPT.
The reason is that GeoPix learns image captioning capabilities from descriptions generated by GPT-4v on VRSBench, whereas the descriptions in RSIEval are manually annotated.
The two differ in description style, content, and length, leading to GeoPix’s suboptimal performance on RSIEval.

\subsection{Visual Grounding}
\label{sec:visual_grounding}
We evaluate our model’s performance on the visual grounding task using the VRSBench~\cite{li2024vrsbench}, which comprises 9,318 images and 16,159 instances, covering both unique and non-unique object referring tasks. Unique objects are the only instance of their category in the image, while non-unique objects have multiple instances, requiring more precise descriptions for accurate localization. 
We report performance in terms of A@0.5 and A@0.7, representing accuracy at IoU thresholds of 0.5 and 0.7, respectively.
As shown in Tab.~\ref{table:visual_grounding}, GeoPix achieves competitive performance on non-unique object referring tasks, with an accuracy of 44.8\% at A@0.5 and 18.2\% at A@0.7. Compared to GeoChat~\cite{kuckreja2023geochat}, GeoPix improves by 0.3\% at A@0.5 and 0.2\% at A@0.7, demonstrating consistent enhancements in our model’s capability for the region-level image understanding task.


\input{tables/ref_vrsbench}

\subsection{Ablation Studies}
We conduct ablation studies on memory capacity, memory fusion strategies, projector strategies, and the proposed two-stage training strategy of GeoPix. These ablation experiments are performed and evaluated on the SIOR-T subset.

\subsubsection{Class-wise Learnable Memory Module}
\begin{figure}[t]
  \centering
  \includegraphics[width=\linewidth]{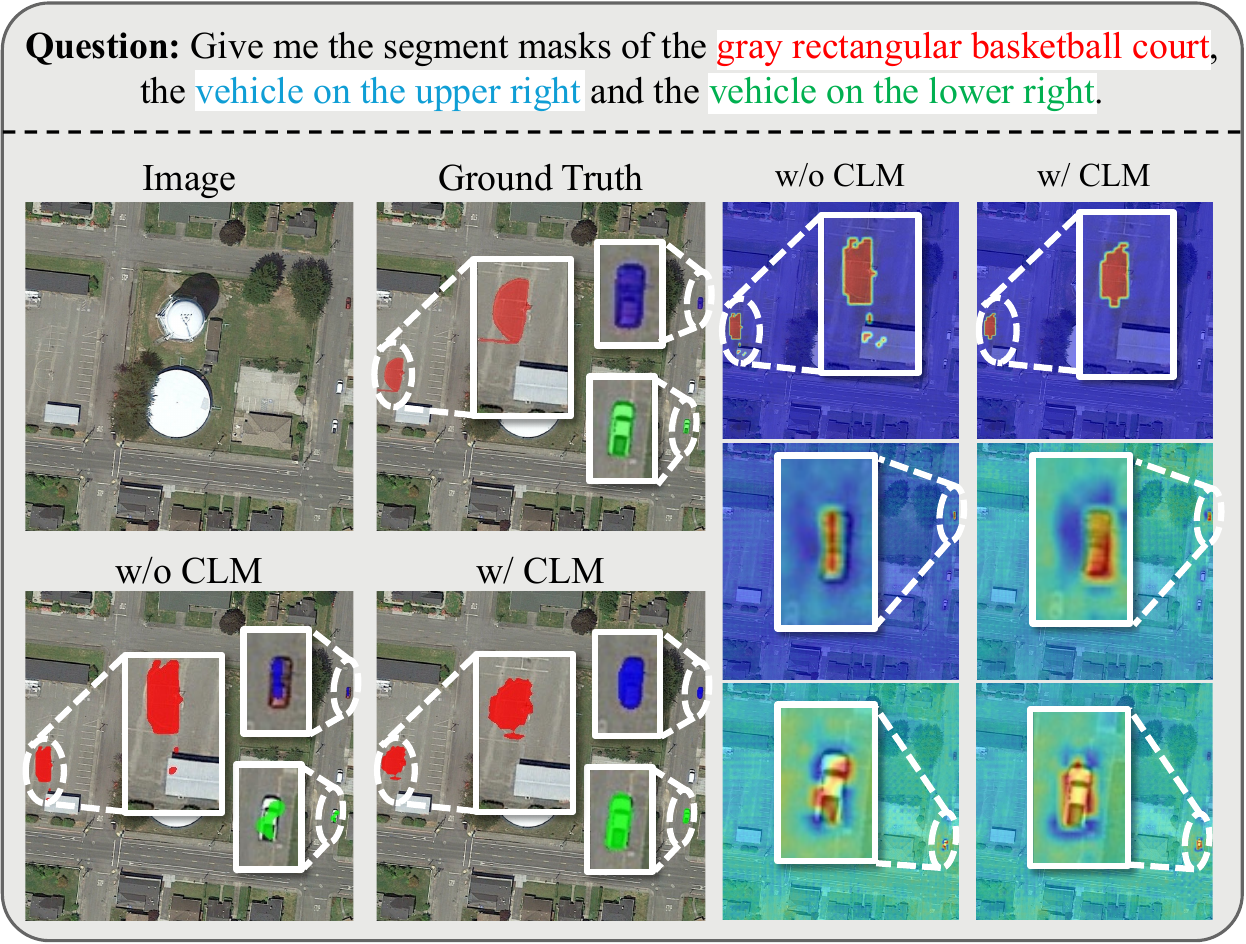}
  \caption{\rz{Comparison between the proposed model with and without the Class-wise Learnable Memory (CLM) module for enhancing the representation of instances in remote sensing images. The post-decoding attention map visualizations highlight how CLM refines the representation of specific instance, leading to more accurate mask predictions.}}
  \label{fig:attnhm}
\end{figure}

\rz{We evaluate the impact of the proposed class-wise learnable memory (CLM) module on segmentation performance. The upper part of Tab.~\ref{table:memory_design} shows the results, where \( N \in \{32, 64, 256\} \) indicates that the model includes the CLM module with a memory capacity of \( N \). Comparing the model without the CLM module and with the CLM module, introducing the CLM module results in a 4-5\% increase in mIoU and cIoU, respectively. Fig.~\ref{fig:attnhm} showcases the post-decoding attention map of the CLM module's output. These visualizations demonstrate the CLM module refines the attention distribution, leading to more accurate mask predictions.
}

\subsubsection{Memory Capacity}
The memory capacity is utilized to enhance the ability of the memory bank to store representative class-wise geo-context. 
We evaluate the impact of memory capacity \( N \in \{32, 64, 256\} \) on segmentation performance. 
The upper part of Tab.~\ref{table:memory_design} shows the results.
When \( N = 64 \), segmentation achieves the highest mIoU. 
Increasing \( N \) to 256 leads to a decrease in mIoU with only marginal improvement in cIoU.
Therefore, in subsequent experiments, we set \( N = 64 \).

\input{tables/alb_design_memory}

\subsubsection{Memory Fusion Strategy}
We then compare the segmentation performance across different memory feature fusion \(\mathcal{F}\) strategies.
Memory feature fusion is adopted to fuse memory features along with the memory capacity \( N \). This process is formulated as \( \mathcal{F}(h^{\ell}) \in \mathbb{R}^{K \times 1 \times D \times H \times W} \). The lower part of Tab.~\ref{table:memory_design} presents the results with a fixed memory capacity of \( N = 64 \).
The Argmax strategy applies an argmax operation over \( N \) to reduce it to 1 in a non-learnable manner.
Conv2D first reshape \( h^{\ell} \) into \( KD \times N \times H \times W \) and employs a 2D convolution.
Conv3D treats \( N \) as the depth dimension and employs a 3D convolution.
The Attention strategy treats \( h^{\ell} \in \mathbb{R}^{HW \times K \times ND} \) as the key and value, while \( \mathcal{I}(m_{\text{init}}^{\ell}) \in \mathbb{R}^{HW \times K \times D} \) serves as the query.
The Dual Attention strategy first uses \( \mathcal{I}(m_{\text{init}}^{\ell}) \) as the key and value and \( h^{\ell} \in \mathbb{R}^{HW \times K \times ND} \) as the query in the first attention step. The roles are then reversed, with \( h^{\ell} \) serving as the key and value, and \( \mathcal{I}(m_{\text{init}}^{\ell}) \) as the query, as formulated in the Attention strategy.
As shown in the lower part of Tab.~\ref{table:memory_design}, the Conv3D strategy achieves the optimal balance between performance and parameter efficiency. Comparing the Argmax strategy with others, the learnable fusion of memory capacity \( N \) enhances mIoU and cIoU by 2\%-4\%. Although the Dual Attention strategy achieves the highest cIoU, its introduction of significantly more parameters results in reduced efficiency.

\input{tables/alb_projector}
\input{tables/alb_training}

\subsubsection{Independent or Shared Projectors}
We next conduct ablation studies on different projector strategies.
The independent projector strategy uses separate rough and detail projectors to process features at different scales, whereas the shared projector strategy applies a single projector for all feature scales.
Tab.~\ref{table:ab_projector_strategies} presents the comparison results, highlighting the superiority of independent projectors over shared ones. The results reveal a 4–5\% improvement in mIoU and cIoU on the SIOR-T subset, confirming that decoupling projection across different visual pathways enhances segmentation performance.

\subsubsection{Two-stage Training Strategies}
We finally conduct ablation studies to analyze the impact of different training strategies on the performance of multiple tasks, including image captioning (IC), visual question answering (VQA), visual grounding (Grd.), and multi-referring segmentation (Seg.).
We compare single-stage and two-stage approaches.
A representative metric is selected for each task to evaluate the results, as shown in Tab.~\ref{table:training_strategy}.
We assess the performance of the models in each stage of the proposed two-stage training strategy, with the training cost measured in A800 GPU hours.
For the single-stage training strategy, we randomly select 8,000 samples from VRSBench and combine them with the SIOR-T subset to form the training set, which is trained for 50 epochs.
As shown in Tab.~\ref{table:training_strategy}, the proposed two-stage training strategy achieves superior performance across all tasks while maintaining the highest efficiency. By adjusting the data ratio in multi-task learning, the two-stage training approach effectively mitigates the tug-of-war problem inherent in multi-task optimization for pixel-level MLLMs, leading to improved performance with reduced computational cost.

\subsection{Limitations and Future Work}
\begin{figure}[t]
  \centering
  \includegraphics[width=\linewidth]{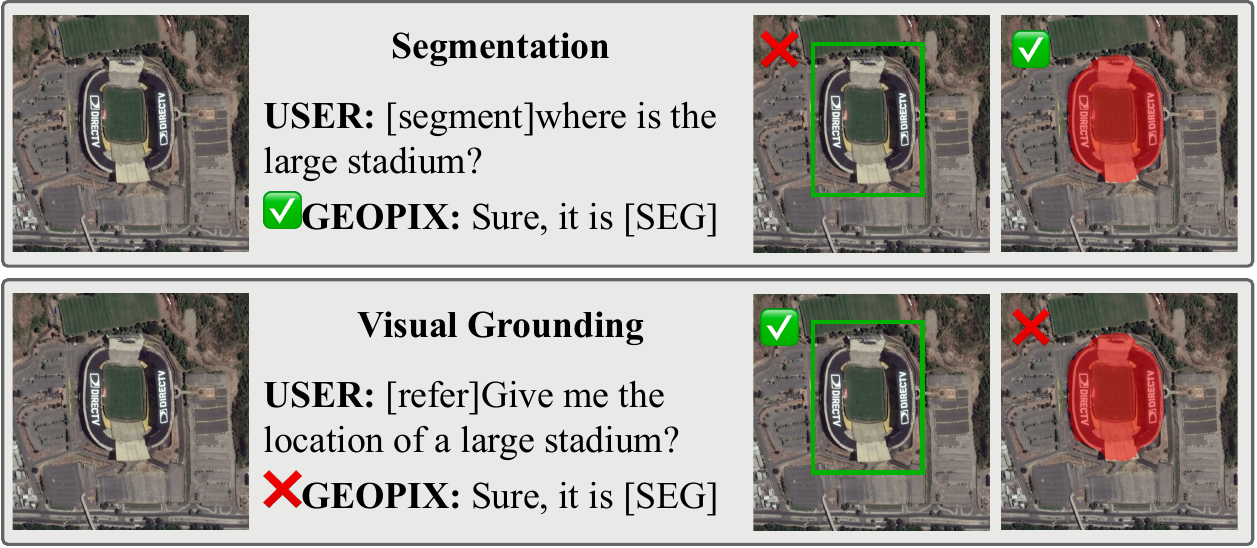}
  \caption{\rz{Failure case of GeoPix: Task confusion causing segmentation tokens to appear in visual grounding results. This leads to the output of the corresponding instance mask when performing visual grounding as specified by the user.}}
  \label{fig:failure_case}
\end{figure}

\rz{GeoPix enables the segmentation of specific instances in remote sensing images based on user instructions, extending beyond previous models that focus on image- and region-level tasks. However, several challenges remain. The challenge arises when user instructions for both referring segmentation and visual grounding tasks inadvertently include queries about the location of instances. As shown in Fig.~\ref{fig:failure_case}, this results in task confusion in GeoPix, leading to incorrect in outputs, with segmentation tokens appearing in visual grounding results. In future work, we will focus on addressing these challenges. We plan to utilize MLLMs to expand and refine instructions in the training data to further mitigate task confusion.

Additionally, due to current resource constraints, we are unable to perform manual verification of GeoPixInstruct. Instead, we leverage vision fine-tuning with human feedback to enhance the accuracy of the generated descriptions. In the future, we intend to leverage MLLMs to enhance the quality of generated descriptions and develop an MLLM-driven filtering pipeline to minimize the reliance on manual verification. Furthermore, we seek to expand GeoPix’s capabilities to effectively handle temporal tasks, such as change detection and weather prediction.}

\section{Conclusion}
In this paper, we propose GeoPix, a multi-modal large language model (MLLM) for remote sensing that extends the capabilities of previous RS MLLMs from image- and region-level to pixel-level image understanding. 
We design a class-wise learnable memory module to enhance segmentation accuracy by adaptively extracting and storing class-wise regional geo-context. 
To facilitate pixel-level RS MLLM training, we constructed GeoPixInstruct by integrating existing datasets and designing a description generation pipeline powered by GPT-4o to produce instance-level descriptions.
Moreover, we propose a two-stage training strategy to mitigate the tug-of-war issue in multi-modal, multi-task optimization of pixel-level RS MLLMs. 
Overall, GeoPix exhibits superior performance on pixel-level tasks while maintaining competitive capabilities at both the image and region levels.


\printbibliography


\end{document}

%% file: tables/intro_compare_data.tex
\begin{table}[t]
\caption{Comparison between existing remote sensing datasets and our GeoPixInstruct\label{tab:dataset_compare}}
\centering
\resizebox{\linewidth}{!}{%
\begin{tabular}{cc|c|c|ccc}
\hline
\multicolumn{2}{c|}{\multirow{2}{*}{Dataset}}      & \multirow{2}{*}{\#Image} & \multirow{2}{*}{Instruct} & \multicolumn{3}{c}{Instance Annotations} \\
\multicolumn{2}{c|}{}   &    &     & bbox      & description      & mask      \\ \hline
\multicolumn{1}{c|}{\multirow{3}{*}{SAMRS~\cite{SAMRS}}} & SIOR & 23,463  & \ding{55}  & \ding{51}  & \ding{55}     & \ding{51}  \\
\multicolumn{1}{c|}{}  & FAST & 64,148  & \ding{55}  & \ding{51}  & \ding{55}     & \ding{51}  \\
\multicolumn{1}{c|}{}  & SOTA & 17,480  & \ding{55}  & \ding{51}  & \ding{55}     & \ding{51}  \\ \hline
\multicolumn{2}{c|}{RefSegRS~\cite{yuan2024rrsis}}       & 4,420   & \ding{55}  & \ding{55}   & \ding{51}    & \ding{51}  \\ 
\multicolumn{2}{c|}{RRSIS-D~\cite{liu2024rmsin}}  & 17,402  & \ding{55}  & \ding{55}   & \ding{51}    & \ding{51}  \\ \hline
\multicolumn{2}{c|}{RSVGD~\cite{rsvgd}}   & 17,402  & \ding{55}  & \ding{51}  & \ding{51}    & \ding{55}   \\ \hline
\multicolumn{2}{c|}{GeoChatInstruct~\cite{kuckreja2023geochat}}     & 252,818  & \ding{51}  & \ding{55}  & \ding{55}    & \ding{55}   \\
\multicolumn{2}{c|}{VRSBench~\cite{li2024vrsbench}}       & 29,614  & \ding{51}  & \ding{51}  & \ding{51}    & \ding{55}   \\ \hline
\multicolumn{2}{c|}{\textbf{GeoPixInstruct}}     & 65,463  & \ding{51}  & \ding{51}  & \ding{51}    & \ding{51}  \\ \hline
\end{tabular}
}
\end{table}

%% file: tables/data.tex
\begin{table}[t]

\centering
\caption{Statistics of GeoPixInstruct. The size refers to the width and height of square images. \rz{The GSD refers to the ground sampling distance measured in meters.} \(\varphi\) represents the number of instances per image and the \(\theta\) represents the mask coverage ratio within the images.}
\label{table:data}
\resizebox{\linewidth}{!}{%
\begin{tabular}{lccccccc}
\hline
Subset & \#Img. & \#Cat. & \#Inst. & Size  & \rz{GSD (m)} & Avg. \(\varphi\)   & Avg. \(\theta\)           \\ \hline
SIOR-T & 17,402 & 20     & 38,320  & 800   & \rz{0.5-30}  & 2.20              & 4.34                 \\
FAST-T & 38,565 & 37     & 77,276  & 600   & \rz{0.3-0.8} & 2.61              & 1.12                 \\
SOTA-T & 9,038  & 18     & 24,816  & 1024  & \rz{0.1-4.5} & 2.00              & 0.94                 \\ \hline
\end{tabular}
}
\end{table}

%% file: tables/seg_v1.tex
\begin{table*}[t]
\centering
\caption{Comparison of model performance on the multi-referring segmentation in the GeoPixInstruct. \rz{VG refers to Visual Grounding, with the corresponding IoU calculated for the bounding boxes. VG+SAM/SAM2 refers to using the Visual Grounding results as conditions for prediction with SAM/SAM2. $^\dagger$ indicates that the models are trained using the proposed two-stage training strategy. $^*$ indicates that the models are specialized in referring segmentation.}}
\label{table:seg}
\resizebox{\linewidth}{!}{%
\begin{tabular}{l|ccccc|ccccc|ccccc}
\hline
\multirow{3}{*}{Method}  & \multicolumn{5}{c|}{SIOR-T}                           & \multicolumn{5}{c|}{FAST-T}                           & \multicolumn{5}{c}{SOTA-T}                           \\ \cline{2-16} 
                                          &   \rz{\rz{Small}} & \rz{\rz{Medium}} & \rz{\rz{Large}} & \multicolumn{2}{c|}{Overall} & \rz{Small} & \rz{Medium} & \rz{Large} & \multicolumn{2}{c|}{Overall} & \rz{Small} & \rz{Medium} & \rz{Large} & \multicolumn{2}{c}{Overall} \\
                                           & \rz{mIoU}  & \rz{mIoU}   & \rz{mIoU}  & mIoU          & cIoU         & \rz{mIoU}  & \rz{mIoU}   & \rz{mIoU}  & mIoU          & cIoU         & \rz{mIoU}  & \rz{mIoU}   & \rz{mIoU}  & mIoU         & cIoU         \\ \hline
RRSIS$^*$ & \rz{36.98} & \rz{73.57}  & \rz{78.99} & 57.80         & 72.85        & \rz{51.75} & \rz{\textbf{72.63}}  & \rz{77.25} & 49.46         & 57.14        & \rz{21.27} & \rz{51.97}  & \rz{\textbf{75.73}} & 24.30        & 42.89        \\
RMSIN$^*$ & \rz{60.72} & \rz{79.40}  & \rz{86.24} & 65.10         & 78.27        & \rz{\textbf{55.48}} & \rz{65.82}  & \rz{64.90} & \textbf{56.59}         & 56.88        & \rz{\textbf{42.60}} & \rz{55.25}  & \rz{53.42} & \textbf{44.42}        & 45.12        \\ \hline
LISA  & \rz{55.11} & \rz{80.73}  & \rz{85.38} & 69.48         & 79.61        & \rz{38.73} & \rz{48.10}  & \rz{63.68} & 40.14         & 40.81        & \rz{30.63} & \rz{44.40}  & \rz{66.50} & 32.29        & 37.69        \\
PixelLM & \rz{45.03} & \rz{72.27}  & \rz{77.47} & 60.37         & 72.73        & \rz{34.21} & \rz{47.70}  & \rz{49.16} & 36.05         & 40.61        & \rz{23.42} & \rz{38.37}  & \rz{52.19} & 25.39        & 35.48        \\ \hline
\rz{LISA$^\dagger$}                                & \rz{62.48} & \rz{86.37}  & \rz{95.75} & \rz{82.14}         & \rz{87.38}        & \rz{38.62} & \rz{49.42}  & \rz{75.82} & \rz{41.98}         & \rz{52.05}        & \rz{27.88} & \rz{46.86}  & \rz{68.23} & \rz{31.02}        & \rz{41.88}        \\
\rz{PixelLM$^\dagger$}                          & \rz{66.93} & \rz{85.71}  & \rz{93.28} & \rz{78.05}         & \rz{86.97}        & \rz{40.23} & \rz{52.25}  & \rz{77.28} & \rz{42.34}         & \rz{52.17}        & \rz{29.26} & \rz{47.42}  & \rz{55.83} & \rz{30.12}        & \rz{41.22}        \\ \hline
\rz{VG}        & \rz{33.56} & \rz{66.30}  & \rz{71.96} & \rz{51.40}         & \rz{63.15}        & \rz{30.42} & \rz{52.69}  & \rz{62.95} & \rz{33.08}         & \rz{43.05}        & \rz{22.70} & \rz{56.47}  & \rz{75.19} & \rz{26.35}        & \rz{46.72}        \\ 
\rz{VG+SAM}                 & \rz{40.37} & \rz{79.21}  & \rz{74.38} & \rz{60.58}         & \rz{69.58}        & \rz{37.22} & \rz{57.70}  & \rz{40.15} & \rz{39.36}         & \rz{37.98}        & \rz{25.83} & \rz{57.87}  & \rz{43.70} & \rz{29.36}        & \rz{39.07}        \\
\rz{VG+SAM2}              & \rz{34.30} & \rz{73.36}  & \rz{62.62} & \rz{54.15}         & \rz{62.10}        & \rz{31.69} & \rz{55.04}  & \rz{47.91} & \rz{34.27}         & \rz{37.14}        & \rz{21.11} & \rz{53.18}  & \rz{38.60} & \rz{24.40}        & \rz{33.27}      \\ 

\textbf{GeoPix} & \rz{\textbf{75.49}} & \rz{\textbf{92.37}}  & \rz{\textbf{96.27}} & \textbf{84.25}         & \textbf{89.82}        & \rz{50.73} & \rz{71.76}  & \rz{\textbf{80.63}} & 53.54         & \textbf{57.25}        & \rz{32.43} & \rz{\textbf{58.01}}  & \rz{59.88} & 35.37        & \textbf{45.24}        \\ \hline
\end{tabular}
}
\end{table*}

%% file: tables/vqa_vrsbench.tex
\begin{table*}[t]
\centering
\caption{Comparison of model performance on the visual question answering task in the VRSBench. All accuracy scores are evaluated by GPT-4 and reported as percentages.}
\label{table:vqa_vrsbench}
\begin{tabularx}{\textwidth}{l|>{\centering\arraybackslash}X>{\centering\arraybackslash}X>{\centering\arraybackslash}X>{\centering\arraybackslash}X>{\centering\arraybackslash}X>{\centering\arraybackslash}X>{\centering\arraybackslash}X>{\centering\arraybackslash}X>{\centering\arraybackslash}X>{\centering\arraybackslash}X>{\centering\arraybackslash}X}
\hline
Method  & Category & Presence & Quantity & Color & Shape & Size & Position & Direction & Scene & Reasoning & Avg. \\ \hline
MiniGPT4v2  & 46.2  & 74.1  & 47.3  & 44.4  & 28.6  & 17.2  & 23.3  & 15.3  & 38.7  & 36.3  & 37.1  \\
LLaVA-1.5  & 62.8  & 89.2  & 50.4  & 57.8  & 58.5  & 52.3  & 56.9  & 50.7  & 66.0  & 64.9  & 60.9  \\
Mini-Gemini & 58.7  & 89.4  & 50.0  & 57.9  & 57.9  & 53.7  & 54.8  & 50.1  & 65.0  & 64.3  & 43.0  \\ \hline
GeoChat  & 60.4  & 89.9  & 47.5  & 58.7  & 59.1  & 52.3  & 57.0  & 50.3  & 66.1  & 64.9  & 60.6  \\ 
\textbf{GeoPix}  & \textbf{84.8}  & \textbf{91.5}  & \textbf{57.5}  & \textbf{68.3}  & \textbf{68.5}  & \textbf{56.9}  & \textbf{68.0}  & \textbf{53.0}  & \textbf{81.9}  & \textbf{71.8}  & \textbf{74.9}  \\ \hline
\end{tabularx}
\end{table*}

%% file: tables/vqa_rsieval.tex
\begin{table*}[t]
\centering
\caption{Comparison of model performance on the visual question answering task in the RSIEval. GeoPix is evaluated in a zero-shot manner on the RSIEval dataset. all accuracy scores are evaluated by GPT-4 and reported as percentages. $^\dagger$ indicates models without fine-tuning in RSICap.}
\label{table:vqa_rsieval}
\begin{tabularx}{\textwidth}{l|ccccccccccc}
\hline
Method  & Presence & Quantity & Color & Absolute pos. & Relative pos. & Area comp. & Road dir. & Image & Scene & Reasoning & Avg.  \\ \hline
BLIP2$^\dagger$   & 60.41  & 26.02  & 43.24 & 7.69  & 13.16  & 58.14  & 33.33  & 74.42 & 43.24 & 47.50  & 45.56 \\
MiniGPT4$^\dagger$   & 29.70  & 9.76  & 31.53 & 1.54  & 1.32  & 16.28  & 0.00  & 34.88 & 24.32 & 17.50  & 21.82 \\ \hline

BLIP2   & 39.74  & 7.50  & 25.86 & 4.40  & 2.50  & 64.58  & 60.00  & 45.83 & 8.89  & 29.09  & 32.04 \\
MiniGPT4   & 59.22  & 15.83  & 37.93 & 20.88  & 22.50  & 27.08  & 0.00  & 60.42 & 73.33 & 23.64  & 37.87 \\ \hline
RSGPT   & \textbf{81.22}  & \textbf{39.02}  & 54.05 & 38.46  & \textbf{35.53}  & 62.79  & \textbf{66.67}  & \textbf{93.02} & \textbf{89.19} & \textbf{70.00}  & 65.24 \\
\textbf{GeoPix$^\dagger$}  & 77.20  & 37.81  & \textbf{56.89}  & \textbf{47.25}  & 15.00  & \textbf{75.00}  & 40.00  & 86.45  & 86.67  & 60.00  & \textbf{65.53}  \\ \hline
\end{tabularx}
\end{table*}

%% file: tables/cap_rsieval.tex
\begin{table*}[t]
\centering
\caption{Comparison of model performance on the image captioning task in the RSIEval. All scores are evaluated and reported as percentages. $^\dagger$ indicates models w/o fine-tuning on rsicap.}
\label{table:cap_rsieval}
\begin{tabularx}{\textwidth}{l|>{\centering\arraybackslash}X>{\centering\arraybackslash}X>{\centering\arraybackslash}X>{\centering\arraybackslash}X>{\centering\arraybackslash}X>{\centering\arraybackslash}X>{\centering\arraybackslash}X>{\centering\arraybackslash}X}
\hline
Method & BLEU-1 & BLEU-2 & BLEU-3 & BLEU-4 & METEOR & ROUGE\_L & CIDEr & CLAIR \\ \hline
BLIP2  & 0.15  & 0.07  & 0.03  & 0.01  & 3.40  & 11.57  & 0.09  & 9.20  \\
MiniGPT4v1  & 47.89  & 33.46  & 23.71  & 17.26  & 19.87  & 35.32  & 16.25  & 26.35  \\ \hline
RSGPT  & \textbf{47.85}  & \textbf{36.06}  & \textbf{27.97}  & \textbf{22.07}  & \textbf{23.58}  & \textbf{41.73}  & \textbf{31.35}  & 44.80  \\ 
\textbf{GeoPix}$^\dagger$  & 18.66  & 9.59  & 5.30  & 2.76  & 9.36  & 16.12  & 2.86 & 19.11  \\ \hline
\end{tabularx}
\end{table*}

%% file: tables/cap_vrsbench.tex
\begin{table*}[t]
\centering
\caption{Comparison of model performance on the image captioning task in the VRSBench. All scores are reported as percentages. Bold indicates the best overall performance, while underlined denotes the best result among remote sensing domain-specific models.}
\label{table:cap_vrsbench}
\begin{tabularx}{\textwidth}{l|>{\centering\arraybackslash}X>{\centering\arraybackslash}X>{\centering\arraybackslash}X>{\centering\arraybackslash}X>{\centering\arraybackslash}X>{\centering\arraybackslash}X>{\centering\arraybackslash}X>{\centering\arraybackslash}X}
\hline
Method  & BLEU-1 & BLEU-2 & BLEU-3 & BLEU-4 & METEOR & ROUGE\_L & CIDEr & CLAIR  \\ \hline
MiniGPT4v2  & 36.8  & 22.4  & 13.9  & 8.7  & 17.1  & 30.8  & 21.4  & 73   \\
LLaVA-1.5  & \textbf{48.1}  & \textbf{31.5}  & \textbf{21.2}  & \textbf{14.7}  & 21.9  & \textbf{36.9}  & \textbf{33.9}  & \textbf{78}  \\
Mini-Gemini & 47.6  & 31.1  & 20.9  & 14.3  & 21.5  & 36.8  & 33.5  & 77  \\ \hline
GeoChat  & \underline{46.7}  & 30.2  & 20.1  & 13.8  & 21.1  & 35.2  & 28.2  & \underline{77}  \\ 
\textbf{GeoPix}  & \underline{46.7}  & \underline{30.5}  & \underline{20.6}  & \underline{14.0}  & \underline{\textbf{23.4}}  & \underline{36.3}  & \underline{31.3}  & \underline{77}   \\ \hline
\end{tabularx}
\end{table*}

%% file: tables/ref_vrsbench.tex

\begin{table}[t]
\centering
\caption{Comparison of model performance on the visual grounding task in the VRSBench. A@0.5 represents accuracy for predictions with IoU over 0.5, while A@0.7 denotes accuracy at a stricter IoU threshold of 0.7.}
\label{table:visual_grounding}
\resizebox{\linewidth}{!}{%
\begin{tabular}{l|cc|cc|cc}
\hline
\multirow{2}{*}{Method}  & \multicolumn{2}{c|}{Unique}  & \multicolumn{2}{c|}{Non-Unique}  & \multicolumn{2}{c}{All} \\
\cline{2-7}
  &A@0.5  &A@0.7  &A@0.5  &A@0.7  &A@0.5  &A@0.7 \\ \hline
MiniGPT4v2  & 40.7  & 18.9  & 32.4  & 15.2  & 35.8  & 16.8  \\
LLaVA-1.5  & 51.1  & 16.4  & 34.8  & 11.5   & 41.6  & 13.6  \\
Mini-Gemini  & 41.1  & 9.6  & 22.3  & 4.9   & 30.1  & 6.8  \\ \hline
GeoChat  & \textbf{57.4}  & 22.6  & 44.5  & 18.0  & \textbf{49.8}  & 19.9  \\ 
\textbf{GeoPix}  &57.0  & \textbf{22.7}  &\textbf{44.8}  &\textbf{18.2}  & \textbf{49.8}  & \textbf{20.0}  \\ \hline
\end{tabular}
}
\end{table}

%% file: tables/alb_design_memory.tex
\begin{table}[t]
\centering
\caption{Comparison of memory capacity and memory fusion. The upper part of the table compares different memory capacities. The lower part evaluates the performance of various memory fusion methods under a fixed capacity.}
\label{table:memory_design}
\begin{tabularx}{\linewidth}{l|c|>{\centering\arraybackslash}X>{\centering\arraybackslash}X}
\hline
\multirow{2}{*}{Design of CLM module} & \multirow{2}{*}{Learnable Parameter}  & \multicolumn{2}{c}{SIOR-T} \\ \cline{3-4} 
    &    & mIoU       & cIoU      \\ \hline
\multicolumn{4}{c}{\textit{Memory Capacity \( N\) (Fusion by Conv3D)}} \\ \hline
\rz{w/o CLM module}    &  0  & 76.56      & 83.25     \\
32     &  2.7K  & 80.77      & 88.12     \\
64     &  2.7K  & \textbf{84.25}  & 89.82     \\
256    &  2.7K  & 83.88      & \textbf{89.99}      \\ \hline
\multicolumn{4}{c}{\textit{Memory Fusion (Capacity = 64)}} \\ \hline
Argmax   &  0  &  79.24  & 87.21     \\
Conv2D   &  2.7K    &  83.88  & 89.42     \\
Conv3D   &  2.7K    &  \textbf{84.25}  & 89.82      \\ 
Attention       &  532.8K  &  83.91  & 88.92     \\
Dual Attention       &  34.6M   &  84.20  & \textbf{89.95}      \\ \hline
\end{tabularx}
\end{table}

%% file: tables/alb_projector.tex
\begin{table}[t]
\centering
\caption{comparison of different projector strategies.}
\label{table:ab_projector_strategies}
\begin{tabular}{l|cc}
\hline
\multirow{2}{*}{Projector strategy} & \multicolumn{2}{c}{SIOR-T} \\ \cline{2-3}
     & mIoU  & cIoU   \\ \hline
Shared    & 74.65       & 85.31  \\
Independent    & \textbf{84.25}   & \textbf{89.82}  \\ \hline
\end{tabular}
\end{table}

%% file: tables/alb_training.tex
\begin{table}[t]
\centering
\caption{comparison of different training strategies in terms of computational cost and performance on multi-modal tasks.}
\label{table:training_strategy}
\begin{tabular}{l|c|cccc}
\hline
Training   & \multirow{2}{*}{Cost} & IC             & VQA           & Grd.            & Seg.                 \\ \cline{3-6} 
strategy   &                       & CLAIR          & Avg.          & All.A@0.5       & SIOR-T cIoU          \\ \hline
Single-stage & 1,010                 & 71             & 73.8          & 39.5            & 85.94                \\
Stage1     & 101                   & 72             & 74.8          & 47.4            & 74.73                \\
+ Stage2   & 125                   & \textbf{77}    & \textbf{74.9} & \textbf{49.8}   & \textbf{89.95}       \\ \hline
\end{tabular}
\end{table}